# VLSI Architectures of Forward Kinematic Processor for Robotics Applications

S. Roy, S Paul, and T. K. Maiti, *Member, IEEE*

*Abstract*— This paper aims to get a comprehensive review of current-day robotic computation technologies at VLSI architecture level. We studied several repots in the domain of robotic processor architecture. In this work, we focused on the forward kinematics architectures which consider CORDIC algorithms, VLSI circuits of WE DSP16 chip, parallel processing and pipelined architecture, and lookup table formula and FPGA processor. This study gives us an understanding of different implementation methods for forward kinematics. Our goal is to develop a forward kinematics processor with FPGA for real-time applications, requires a fast response time and low latency of these devices, useful for industrial automation where the processing speed plays a great role.

*Keywords*: VLSI Architecture, Kinematic, Robotics, FPGA, High-Speed Computation.

## I. INTRODUCTION

In the dynamic landscape of robotics, achieving real-time and precise kinematics calculations remains a formidable challenge. This paper introduces a comprehensive study of cutting-edge methodologies and architectures devised to tackle the complexities of multi-degree-of-freedom manipulators. The collective efforts outlined in these studies aim to overcome the inherent computational hurdles associated with intricate robotic systems by using diverse approaches such as CORDIC algorithms, VLSI architectures, and FPGA implementations. Efficiency, modularity, and real-time processing are investigated using pipelined designs, homogeneous link transformation matrices, and parallel processing. Robotic technology and industrial automation could benefit from searching for the best kinematics computation techniques, which are becoming increasingly important as technology develops.

Zheng *et. al.*, implemented the FPGA-based CORDIC algorithm, which enhances real-time manipulator precision with parallel processing for kinematics calculations, addressing challenges in single-processor setups [1]. Lee *et. al.*, introduces a pipelined CORDIC-based architecture for manipulator kinematics, employing homogeneous link transformation matrices and modular CORDIC processors [2], [3]. The flexible approach achieves precision and scalability, demonstrated with a 6-link PUMA robot using 24 CORDIC processors, providing accurate results with a total computation time $(80n + 120)$ $\mu s$. Seshadri *et. al.*, implemented real-time kinematics algorithm on a signal processor using sinusoidal functions and fixed-point calculations, achieving three orders of magnitude speed improvement [4]. Kim *et. al.*, proposes efficient chip architectures for robotics, focusing on a 6-link robot [5]. Utilizing augmented CORDIC processing and a fully-pipelined technique, it optimizes kinematic computations for VLSI implementation. The study explores CORDIC techniques, macro-PE structures, and the Constant-Factor-Redundant CORDIC (CFR-CORDIC) scheme for cost-effective and efficient kinematic calculations. Steven *et. al.*, proposes FPGA-based forward kinematics for the Utah MIT Dexterous Hand, enhancing portability and integration into prosthetics [6]. The research demonstrates practicality and efficiency, despite FPGA constraints. However, development of CORDIC based forward kinematic (FK) takes longer in implementation. Therefore, we proposed the lookup table based FK development and implementation.

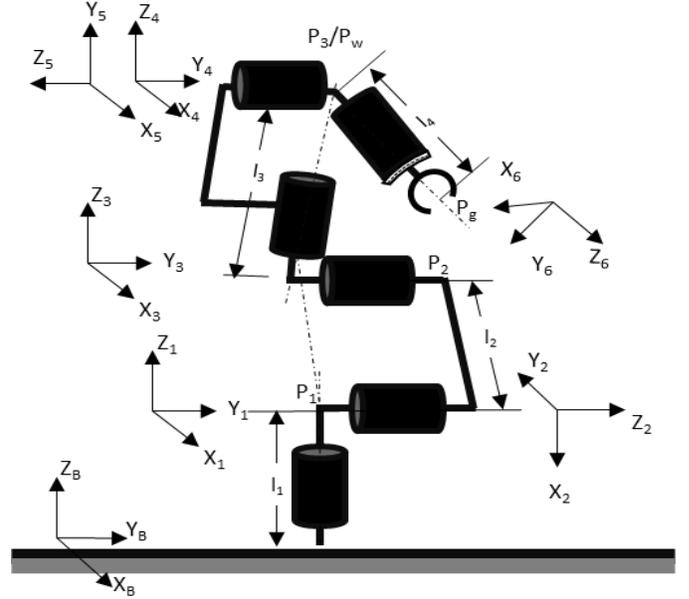

Fig. 1. Schematic diagram of a 6-DOF robotic manipulator in 3D space which consists of several links and joints. Here link $l$ = 1, 2,……,4, three modular 2-DOF rotary joints—$P_1$, $P_2$, and $P_3$/$P_W$, a gripper ($P_g$).

The work has been supported by Science and Engineering Research Board (SERB), Department of Science & Technology, Government of India.

S. Roy, S. Paul, and T. K. Maiti are with the DA-IICT, Reliance Cross Rd, Gandhinagar, Gujarat, 382007 (e-mail: sourav_roy; subhadeep_paul; tapas_kumar@daiict.ac.in).

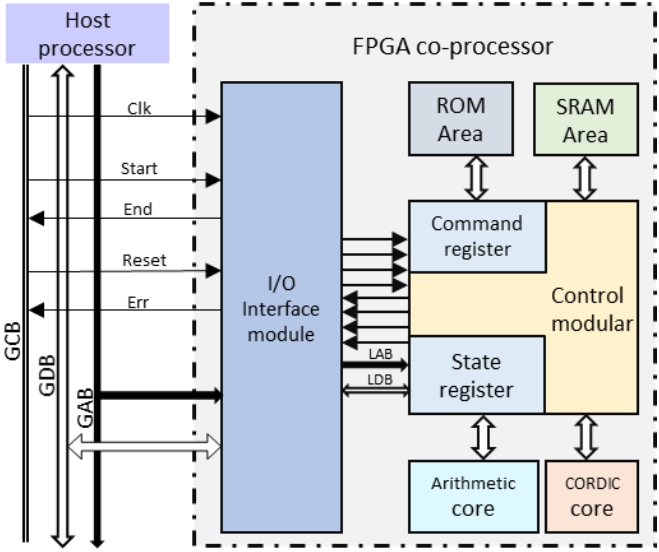

Fig. 2. The architecture of parallel processing system.

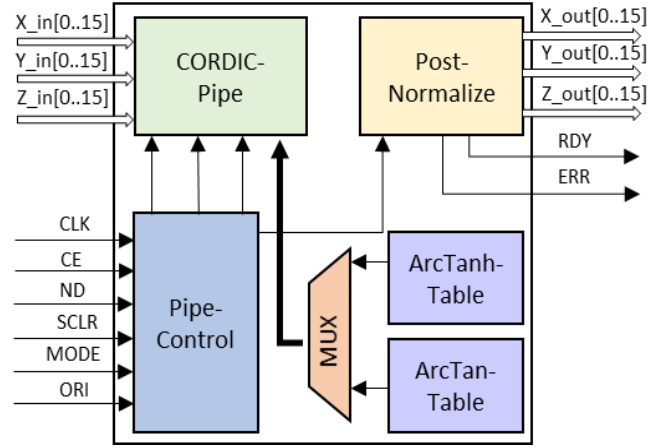

Fig. 3. The architecture of the CORDIC Core structure.

## II. KINEMATIC FOR ROBOTIC MANIPULATOR

Kinematics calculations take a long time, it is challenging to obtain real-time motion for a single DOF on a single processor [1]. The CORDIC approach reduces the amount of time needed to calculate the results of kinematics calculations and also provides exact control over DOF by leveraging pipeline design in parallel processing. Utilizing parallel computing and the CORDIC algorithm, the manipulator's inverse kinematics equations were resolved. Three inputs are required for the CORDIC algorithm: the vector's coordinate components ($X_0$, $Y_0$) and the vector's rotation angle ($Z_0 = \theta$). This study uses three different versions of the CORDIC algorithm. This work lacks an understanding of the CORDIC algorithm's modes and parameters. All of the inverse kinematics equations that the CORDIC cores computed worked with multipliers, adders, and inverters; addition and multiplication operations could be completed using standard arithmetic. The manipulator's inverse kinematics computation is carried out via an FPGA-based CORDIC pipeline, which is a co-processor connected to a host CPU and used to obtain the angle data needed to carry out the kinematics calculation. Five basic building blocks make up the CORDIC core's internal logic. This work uses VHDL code to implement kinematics calculation in an FPGA XC3S2000. The end-effector's perfect precision satisfies the criteria. This technique is typically applicable to real-time manipulator activities in the field or industry with high requirements.

### A. Implementation of Direct Kinematics using CORDIC

In this session we describe a pipelined architecture based on CORDIC that uses a homogeneous link transformation matrix to calculate direct kinematic position solutions. The architecture can be expanded to an n-link manipulator with n 2-stage CORDIC computational modules, requiring a total computation time of ($80n + 120$) $\mu s$ for the position and orientation of the end-effector. Efficient computation of 4×4 homogeneous link transformation matrix $^{i-1}A_i$ using CORDIC processors resulted in a pipelined architecture with an initial delay time of 80 μs. The $^{i-1}A_i$ matrix represents the relationship of a point ($P_i$) in homogeneous coordinates to the $(i-1)^{th}$ coordinate system which is represented by the homogeneous transformation matrix $^0T_i$, which specifies its position and orientation as $P_{i-1} = {}^{i-1}A_i P_i$, here $P_{i-1} = (x_{i-1}, y_{i-1}, z_{i-1}, 1)^T$, $P_i = (x_i, y_i, z_i, 1)^T$,

$$^{i-1}A_i = \begin{cases} \begin{bmatrix} \cos\theta_i & -\cos\alpha_i \sin\theta_i & \sin\alpha_i \sin\theta_i & a_i \cos\theta_i \\ \sin\theta_i & -\cos\alpha_i \sin\theta_i & -\sin\alpha_i \cos\theta_i & a_i \sin\theta_i \\ 0 & \sin\alpha_i & \cos\alpha_i & d_i \\ 0 & 0 & 0 & 1 \end{bmatrix} & \text{for a rotary joint} \\ \begin{bmatrix} \cos\theta_i & -\cos\alpha_i \sin\theta_i & \sin\alpha_i \sin\theta_i & 0 \\ \sin\theta_i & \cos\alpha_i \cos\theta_i & -\sin\alpha_i \cos\theta_i & 0 \\ 0 & \sin\alpha_i & \cos\alpha_i & d_i \\ 0 & 0 & 0 & 1 \end{bmatrix} & \text{for a prismatic joint} \end{cases}$$

$$^0T_i = {}^0A_1 {}^1A_2 \ldots {}^{i-1}A_i = \prod_{j=1}^{i} {}^{j-1}A_j = \begin{bmatrix} x_i & y_i & z_i & P_i \\ 0 & 0 & 0 & 1 \end{bmatrix}$$ for $i = 1, 2, \ldots, n$. In case of PUMA robot, arm equation expressed as [2]

$$^0T_i = \begin{bmatrix} n_x & s_x & a_x & p_x \\ n_y & s_y & a_y & p_y \\ n_z & s_z & a_z & p_z \\ 0 & 0 & 0 & 1 \end{bmatrix} \quad (1)$$

Here,
$n_x = C_1[C_{23}(C_4C_5C_6 - S_4S_6) - S_{23}S_5C_6] - S_1[S_4C_5C_6 + C_4S_6]$
$n_y = S_1[C_{23}\{C_4C_3C_3 - S_4S_6\} - S_{23}S_5C_6] + C_1[S_4C_5C_6 + C_4S_6]$
$n_z = -S_{23}[C_4C_3C_6 - S_4S_6] - C_{23}S_5C_6$

$s_x = C_1[-C_{23}(C_4C_5S_6 + S_4C_6) + S_{23}S_5S_6) - S_1[-S_4C_5S_6 + C_4C_6]$
$s_y = S_1[-C_{23}(C_4C_5S_6 + S_4C_6) + S_{23}S_5S_6 + C_1[-S_4C_3S_6 + C_4C_6]$
$s_z = S_{23}[C_4C_5C_6 - S_4C_6] - C_{23}S_5S_6$

$a_x = C_1(C_{23}C_4C_5 + S_{23}C_5) - S_1S_4S_5$
$a_Y = S_1(C_{23}C_4C_5 + S_{23}C_5) - S_1S_4S_5$
$a_Z = -S_{23}C_4S_5 + C_{23}C_5$

$p_x = C_1[d_6(C_{23}C_4S_5 + S_{23}C_5) + S_{23}d_4 + a_3C_{23} + a_2C_2] - S_1(d_6S_4S_5 + d_2)$
$p_y = S_1[d_6(C_{23}C_4S_5 + S_{23}C_5) + S_{23}d_4 + a_3C_{23} + a_2C_2] - C_1(d_6S_4S_5 + d_2)$
$p_z = d_6(C_{23}C_5 - S_{23}C_4S_5) + S_{23}d_4 - a_3S_{23} - a_2S_2$

$d_i$ and $a_i$ are known PUMA'S link parameters, and $C_i \equiv cos\theta_i$, $S_i \equiv sin\theta_i$, $C_{ij} \equiv cos(\theta_i + \theta_j)$, and $S_{ij} \equiv sin(\theta_i + \theta_j)$. This flexible and modular approach addresses the limitations of current table look-up techniques. By decomposing computations into CORDIC computational modules (CCM), this architecture analyzes computational flow and data dependency by focusing on the chain product of the homogeneous link transformation matrix. Matrix $^{i-1}A_i$ can be decomposed into a product of four basic homogeneous translation/ rotation matrices as $^{i-1}A_i = Tran(z_{i-1}, d_i) Rot(z_{i-1}, \theta_i) Tran(x_i, a_i) Rot(x_i, \alpha_i)$. Here

$$^{i-1}A_i = \begin{bmatrix} 1 & 0 & 0 & 0 \\ 0 & 1 & 0 & 0 \\ 0 & 0 & 1 & 0 \\ 0 & 0 & 0 & 1 \end{bmatrix} \begin{bmatrix} cos\theta_i & -sin\theta_i & 0 & 0 \\ sin\theta_i & cos\theta_i & 0 & 0 \\ 0 & 0 & 1 & 0 \\ 0 & 0 & 0 & 1 \end{bmatrix} * \begin{bmatrix} 1 & 0 & 0 & a_i \\ 0 & 1 & 0 & 0 \\ 0 & 0 & 1 & 0 \\ 0 & 0 & 0 & 1 \end{bmatrix} \begin{bmatrix} 1 & 0 & 0 & 0 \\ 0 & cos\alpha_i & -sin\alpha_i & 0 \\ 0 & sin\alpha_i & cos\alpha_i & 0 \\ 0 & 0 & 0 & 1 \end{bmatrix}$$

$$^{i-1}A_i = \begin{bmatrix} cos\theta_i & -sin\theta_i & 0 & 0 \\ sin\theta_i & cos\theta_i & 0 & 0 \\ 0 & 0 & 1 & d_i \\ 0 & 0 & 0 & 1 \end{bmatrix} \begin{bmatrix} 1 & 0 & 0 & a_i \\ 0 & cos\alpha_i & -sin\alpha_i & 0 \\ 0 & sin\alpha_i & cos\alpha_i & 0 \\ 0 & 0 & 0 & 1 \end{bmatrix} \quad (2)$$

Perform a two-step coordinate transformation from a vector in the $i^{th}$ coordinate frame to the same vector in the $(i-1)^{th}$ coordinate frame. The first step is transforming a vector $P_i = (x_i, y_i, z_i, 1)^T$ in the $i^{th}$ coordinate frame into an intermediate vector $X_i^A = (x_i^A, y_i^A, z_i^A, 1)^T$

$$\begin{bmatrix} x_i^A \\ y_i^A \\ z_i^A \\ 1 \end{bmatrix} = \begin{bmatrix} 1 & 0 & 0 & a_i \\ 0 & cos\alpha_i & -sin\alpha_i & 0 \\ 0 & sin\alpha_i & cos\alpha_i & 0 \\ 0 & 0 & 0 & 1 \end{bmatrix} * \begin{bmatrix} x_i \\ y_i \\ z_i \\ 1 \end{bmatrix} = \begin{bmatrix} x_i + a_i \\ y_i C\alpha_i - z_i S\alpha_i \\ z_i C\alpha_i + y_i S\alpha_i \\ 1 \end{bmatrix} \quad (3)$$

Then, the intermediate vector $X_i^A$ is mapped to the desired vector $P_{i-1} = (x_{i-1}, y_{i-1}, z_{i-1}, 1)^T$ in the second step,

$$\begin{bmatrix} x_{i-1} \\ y_{i-1} \\ z_{i-1} \\ 1 \end{bmatrix} = \begin{bmatrix} cos\theta_i & -sin\theta_i & 0 & 0 \\ sin\theta_i & cos\theta_i & 0 & 0 \\ 0 & 0 & 1 & d_i \\ 0 & 0 & 0 & 1 \end{bmatrix} * \begin{bmatrix} x_i^A \\ y_i^A \\ z_i^A \\ 1 \end{bmatrix} = \begin{bmatrix} x_i^A C\theta_i - y_i^A S\theta_i \\ y_i^A C\theta_i + x_i^A S\theta_i \\ z_i^A + d_i \\ 1 \end{bmatrix} \quad (4)$$

In Fig. 4, the CORDIC processor mode m is set to -1, 0, or 1, and $x_0$, $y_0$, and $z_0$ are three inputs, and $x_n$, $y_n$, and $z_n$ are three outputs. In (3) and (4), two CORDIC processors arranged in parallel and functioned briefly described in Step-1 and Step-2,

Step 1-a: CORDIC Processor: CIRC1

$$Input = \begin{cases} x_0 = y_i \\ y_0 = z_i \\ z_0 = \alpha_i \end{cases} \quad Output = \begin{cases} x1_n = y_i C\alpha_i - z_i S\alpha_i \equiv y_i^A \\ y1_n = z_i C\alpha_i - y_i S\alpha_i \equiv z_i^A \\ z1_n = \text{Not used} \end{cases} \quad (5)$$

Step 1-b: CORDIC Processor: LIN1

$$Input = \begin{cases} x_0 = 1 \\ y_0 = a_i \\ z_0 = x_i \end{cases} \quad Output = \begin{cases} x2_n = \text{Not used} \\ y2_n = x_i + a_i \equiv x_i^A \\ z2_n = \text{Not used} \end{cases} \quad (6)$$

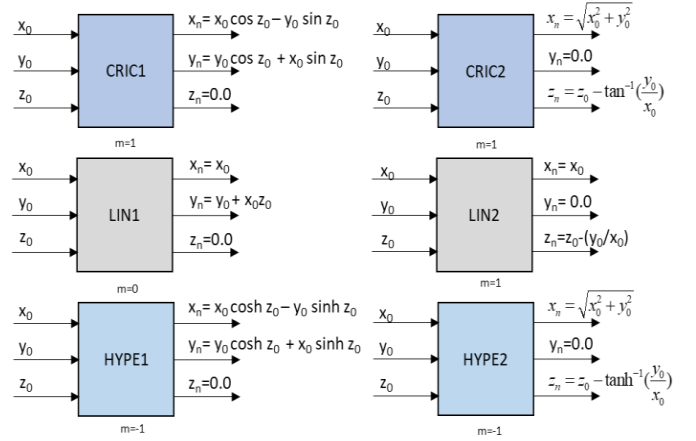

Fig. 4. Elementary functions computed by CORDIC processors.

Step 2-a: CORDIC Processor: CIRC1

$$Input = \begin{cases} x_0 = y2_n \equiv x_i^A \\ y_0 = x1_n \equiv y_i^A \\ z_0 = \theta_i \end{cases} \quad Output = \begin{cases} x3_n = x_i^A C\theta_i - y_i^A S\theta_i \equiv x_{i-1} \\ y3_n = y_i^A C\theta_i - x_i^A S\theta_i \equiv y_{i-1} \\ z3_n = \text{Not used} \end{cases} \quad (7)$$

$x_i^A$ and $y_i^A$ substituted in (3)

$$Output = \begin{cases} x3_n = x_i C\theta_i - y_i C\alpha_i S\theta_i + z_i S\alpha_i S\theta_i + a_i C\theta_{ii} \equiv x_{i-1} \\ y3_n = y_i S\theta_i - y_i C\alpha_i C\theta_i - z_i S\alpha_i C\theta_i + a_i S\theta_{ii} \equiv y_{i-1} \\ z3_n = \text{Not used} \end{cases} \quad (8)$$

Step 2-b: CORDIC Processor: LIN1

$$Input = \begin{cases} x_0 = 1 \\ y_0 = d_i \\ z_0 = y1_i \equiv z_i^A \end{cases} \quad Output = \begin{cases} x4_n = \text{Not used} \\ y4_n = z_i^A + d_i \equiv z_{i-1} \\ z4_n = \text{Not used} \end{cases} \quad (9)$$

Substituted $z_i^A$ in (3)

$$Output = \begin{cases} x4_n = \text{Not used} \\ y4_n = y_i S\alpha_i + z_i C\alpha_i + d_i \equiv z_{i-1} \\ z4_n = \text{Not used} \end{cases} \quad (10)$$

Steps 2-a and 2-b outputs correspond to the matrix-vector multiplication which results end-effector position of robot. Figure 5 displays a CORDIC computational module with a two-stage cascade. The Puma robot arm has used a 6 CORDIC computational module with 24 CORDIC processors which is illustrated in Fig. 3. The CORDIC-based pipeline architecture is flexible, modular, and accurate. It uses four CORDIC processors to form a two-stage cascade CORDIC computational module, providing a modular solution for manipulators with prismatic or rotary joints. The approach is implemented on a CMOS device with 24-bit data processing, converges with an error of $P_4$, and has a total computation time of $(80n + 120)\mu s$ for computing the end-effector's position and orientation.

*B. VLSI Architecture for Direct Kinematics*

A real-time direct kinematics algorithm has been implemented on a signal processor using sinusoidal function generation and fixed-point calculation. The system uses a 36-bit accumulator and a 16*16 multiplier in a parallel architecture. Simulations and hardware execution show a 16-bit resolution, achieving a speed improvement of three orders of magnitude compared to a conventional 16-bit microprocessor. In the Sinusoidal Function Generation, a combination of calculation and table lookup is used for sine/cosine generation. According to the Taylor Series expansion,

$$\sin x = x - x^3\left(\frac{1}{3!}\right) + x^5\left(\frac{1}{5!}\right) - x^7\left(\frac{1}{7!}\right) + \ldots + (-1)^{((r-1)^2)}x^r\left(\frac{1}{r!}\right) + R(\sin)$$

$$\cos x = x - x^2\left(\frac{1}{2!}\right) + x^4\left(\frac{1}{4!}\right) - x^6\left(\frac{1}{6!}\right) + \ldots + (-1)^{r/2}x^{(r-1)}\left(\frac{1}{(r-1)!}\right) + R(\cos)$$

$$|R(\sin)| < x^{(r+1)}/(r+1)!$$

$$|R(\cos)| < x^r / r!$$

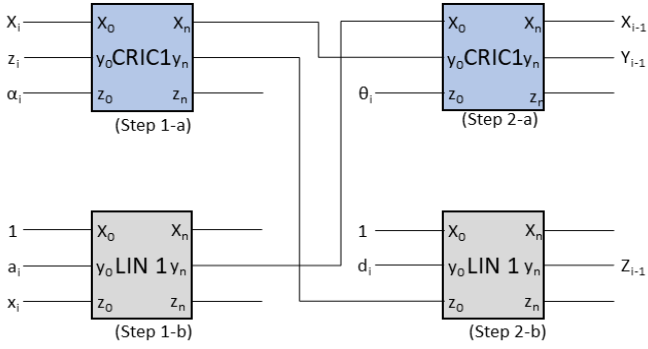

Fig. 5. 2-Stage CORDIC computational module for computing $^{i-1}A_i$.

This approach uses 8 terms for better accuracy than 16-bit words, with internal calculations using 16-bit fixed-point numbers. The algorithm, implemented on the WEtmDSP16, a parallel, pipelined microprocessor with a 60 ns instruction cycle time, is used to implement the method. It has 36-bit accumulators, a 16×16 multiplier, a strong instruction set, and an instruction cache. The chip minimizes power drain with a 0.25W power consumption and serial and parallel I/O support. The chip features a 16-bit data bus, 16×16 multiplier, and 36-bit accumulators, reducing data transfer bottlenecks.

Four addressing modes are available with the WE DSP16 chip: immediate, indirect source, indirect postmodified, and indirect destination.

$$y = 0 \times abcd$$
$$y = *R0$$
$$y = *R0++ \quad (11)$$
$$*R0Pz = y$$

13.3 Mbytes/s of fast data transfer to and from external buses is made possible by the WE DSP16, a 16-bit bidirectional interface. It eliminates glue logic and supports DMA transfers from large external RAM. Trajectory planners and artificial intelligence systems use this interface to transfer robot joint position and orientation, essential for making adjustments and decisions. Serial and parallel I/O, the chip minimizes power drain. Serial and parallel I/O, the chip minimizes power drain.

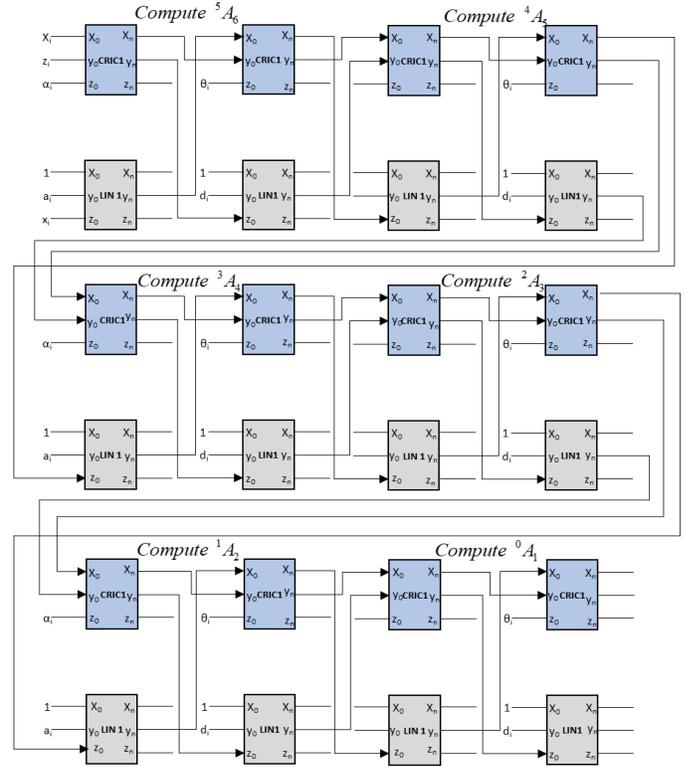

Fig. 6. A CORDIC based pipelined architecture for direct kinematics computation.

*C. Pipelined VLSI Architecture for Robotic Manipulator*

The paper proposes efficient computer chip architectures for computing arm locations and movements in robotics, focusing on a 6-link robot. It utilizes a fully pipelined technique and augmented CORDIC processing elements to optimize kinematic computations. The design is optimized for single-chip VLSI with current MOS technology. The research introduces an augmented CORDIC algorithm that builds on the basic PE and proposes a low-cost direct kinematic computing module. It analyzes traditional CORDIC, redundant CORDIC, and CFR-CORDIC versions. The research also explores a fully-pipelined architecture scheme within the general broad design strategy for robot kinematics processing. CORDIC Techniques. Here the $j^{th}$ joint orientation vector, denoted by $p_j$, is equal to $A_{j*}p_{j-1}$. An intermediate vector, $p_j^A$, is considered between $p_j$ and $p_{j-1}$. Here processor calculates $p_j = Trans\ (w_{j-1}, d_j)*Rot\ (w_{j-1}, \theta_j)p_j^A$ in (stage – 1) and $p_j^A = Trans(x_j, a_j)*Rot\ (x_j, \psi_j)p_{j-1}$ in (stage - 2). The *Trans* (w, d) and *Rot* (w, e) are denoted with block-diagonal matrices that are orthogonally built with two 2×2 matrix transformations and an augmented PE instead of two individual PEs. The expression of $p_j$ derived as

$$p_j \approx \begin{bmatrix} x_j \\ y_j \\ . \\ w_j \\ 1 \end{bmatrix} = \begin{vmatrix} Rot(w_j, \theta_j) & : & 0 \\ . & . & . \\ 0 & : & Trans(w_j, d_j) \end{vmatrix} p_j^A \quad (12)$$

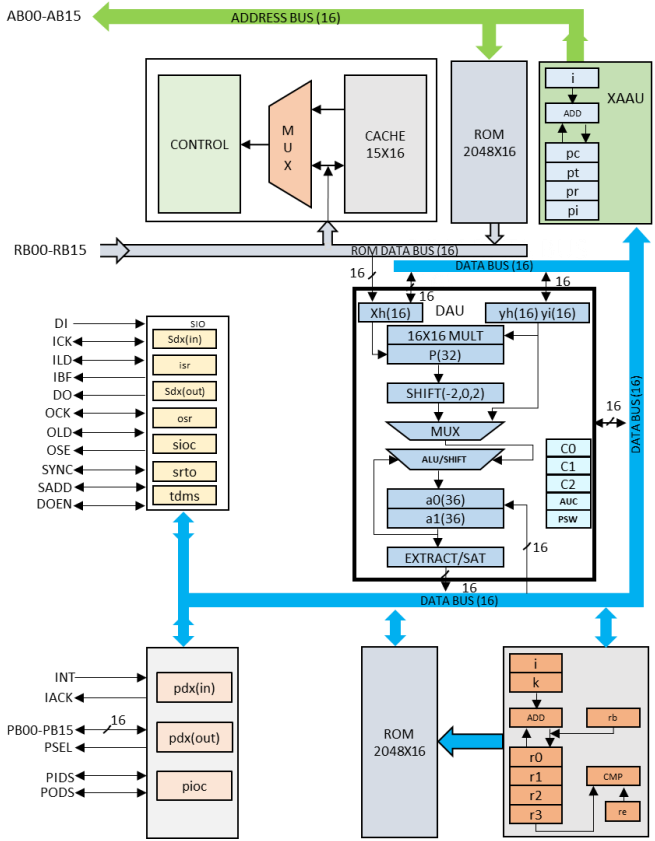

Fig. 7. WE DSP16 Block Diagram.

This section describes a method for describing joint orientations in a robot using vectors ($p_j$), matrix ($A_j$), and an intermediate vector ($p_j^A$). Transformations along each axis are performed using an augmented processing element (PE) and a block-diagonal matrix. This process forms a cascade of stages called macro-PEs, which can be pipelined for efficient computation in multiple joints. Figure 8(a) shows a one-joint processor constructed by cascading two macro-PEs, whereas Fig. 8(b) shows a fully pipelined structure for a six-joint system.

The CORDIC algorithm for macro-PE, separating rotation and translation functions, achieving vector rotation through micro-angle rotations. The Constant-Factor-Redundant CORDIC (CFR-CORDIC) scheme aims to reduce implementation costs by implementing a constant scale factor and dividing micro-iterations into two groups, simplifying the number of correcting iterations and ensuring convergence assurance. The modified recurrences and selection functions for the scheme are described below.

$$X[i+1] = X[i] + \sigma_i 2^{-i} Y[i]$$
$$Y[i+1] = Y[i] - \sigma_i 2^{-i} X[i] \quad (13)$$
$$U[i+1] = 2(U[i] - \sigma_i 2^i \tan^{-1} 2^{-i})$$

Direct kinematics application is exemplified in a processing element (PE) system (Fig. 9(a)), with detailed $X$ and $Y$ recurrence blocks (Fig. 9(b)) using parallel/redundant arithmetic. Figure 9(c) displays Z-recurrence. CFR-CORDIC improves the basic PE by replacing a carry-free adder and affecting sign determination using fractional bits. Micro-pipelined CORDIC (Figure 10) unfolds internal recurrences, achieving speedup in a Z-recurrence micro-PE and a micro-pipelined macro-PE through shifting and recurrence developing for optimal performance. CORDIC techniques for VLSI implementation in direct kinematics are investigated in this paper. The suggested fully parallel macro-PE with redundant arithmetic is cost-effective and shows potential for different and efficient kinematic calculations.

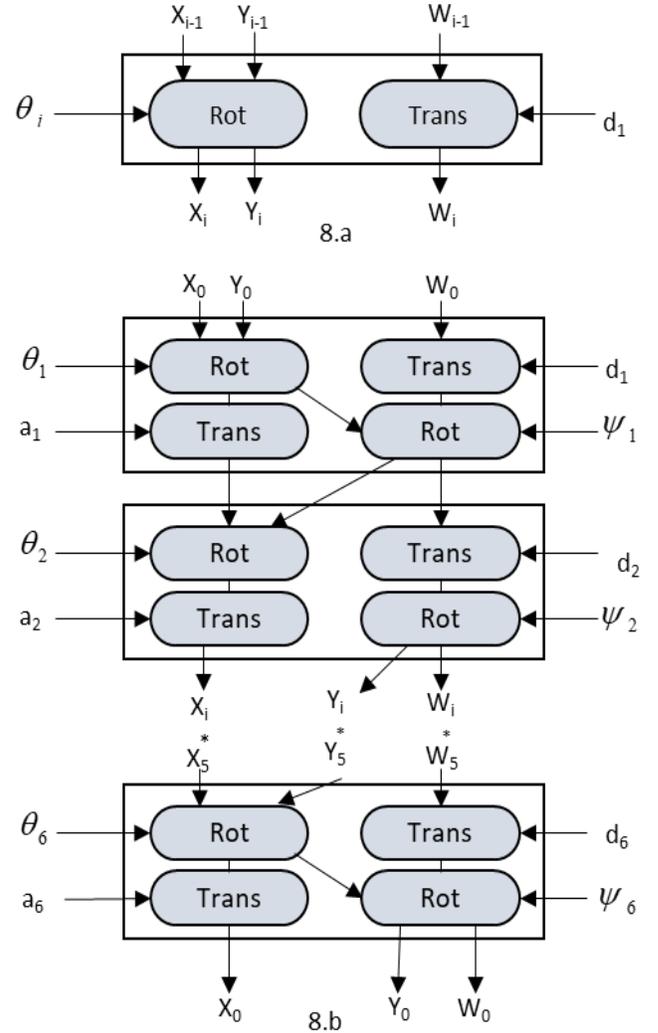

Fig. 8. CORDIC-Basel Pipelined Architecture for Forward Kinematics Computation: (a) a macro-PE, one-stage from orientation to an intermediate; (b) a complete pipelined computation module for a 6-link system.

III FPGA ACCELERATED FORWARD KINEMATICS PROCESSOR FOR ROBOTICS HAND

This session proposes the use of an FPGA processor technique to develop a forward kinematic algorithm for the Utah MIT Dexterous Hand (UMDH). The hardware solution is flexible and dedicated, eliminating the requirement for a real-time operating system and allowing controller integration into portable platforms such as dexterous prosthetic hands in the future. The UMDH forward kinematic algorithm was chosen as an example for using this hardware implementation technique in various robotic systems. The thesis consists of three parts: an examination of the UMDH, an evaluation of the resulting equations, and the development of mathematical, memory storage, and controller functional units using VHDL models. The research demonstrates the practicality and advantages of using FPGA technology for complex robotic algorithms.

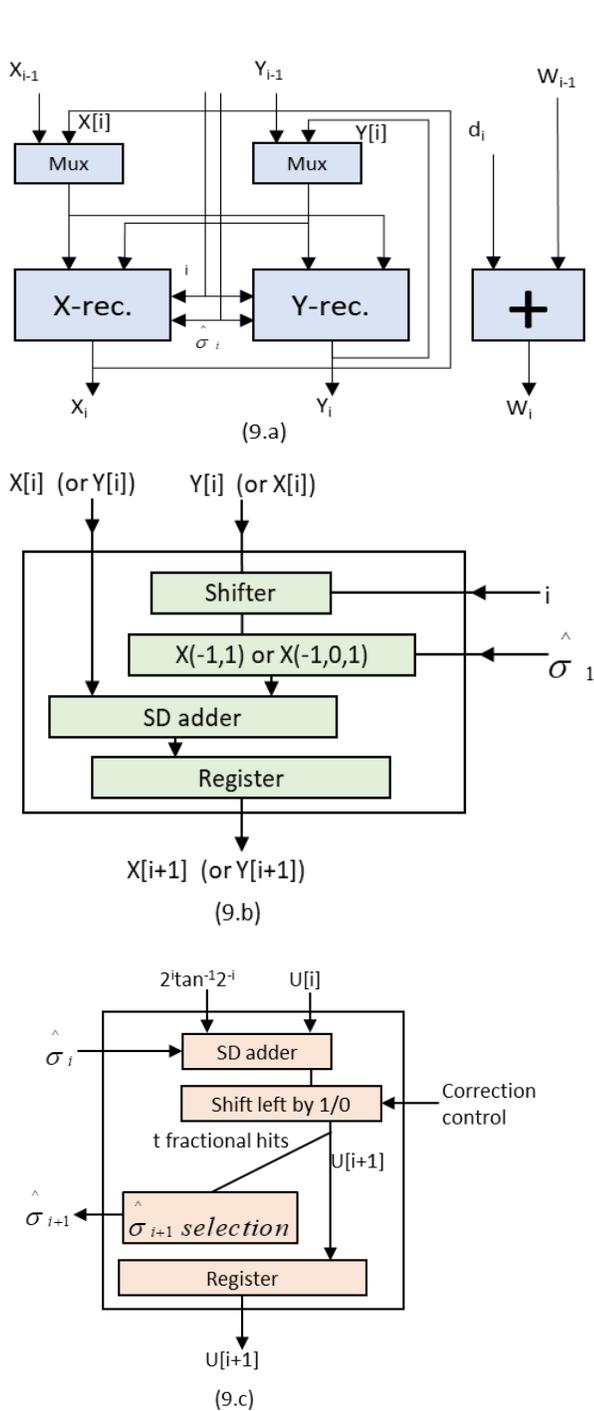

Fig. 9. A parallel/redundant PE: (a) is a macro-PE with $X$- and $Y$-recurrence, (b) details of either block, and (c) is Z-recurrence.

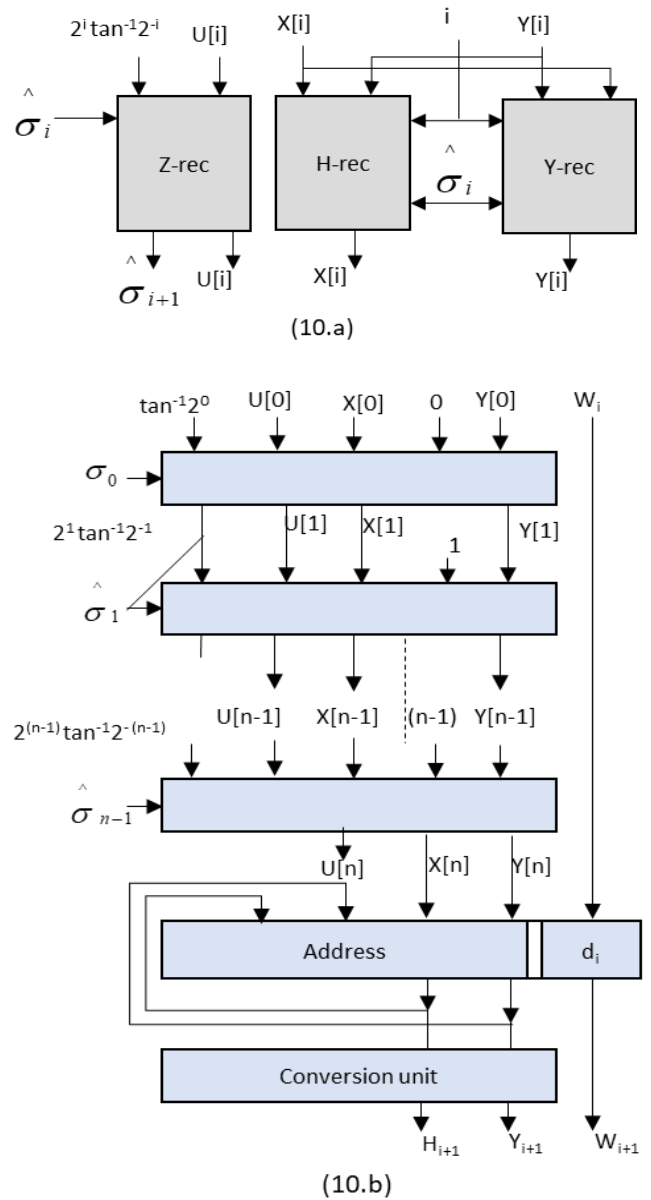

Fig. 10. An n-pipelined redundant PE (a) a micro-PE architecture and (b) $n$-pipelined architecture.

The Utah MIT Dexterous Hand (UMDH) project aims to develop and implement a dedicated forward kinematic processor on a Xilinx Field Programmable Gate Array (FPGA). However, we can use the Altera FPGA in this case. The algorithm is specifically designed for UMDH, where mathematical and transcendental properties are prioritized. Functional units process numeric data efficiently, are integrated into a processing unit, and are synthesized from VHDL code to logic blocks on a Xilinx FPGA. But we can use Verilog code for simplicity. This method simplifies and improves forward kinematic calculations within a specific hardware framework. They investigated a mathematical method for calculating the forward kinematic equations (from Craig, John J. Introduction to Robotics: Mechanics and Control. Reading Ma: Addison-Wesley, 1989) of the thumb mechanism of the Utah MIT

$$^0T_4 = \begin{bmatrix} \cos(\theta_1)\cos(\theta_2+\theta_3+\theta_4) & -\cos(\theta_1)\sin(\theta_2+\theta_3+\theta_4) & \sin(\theta_1) & a_0+\cos(\theta_1)(a_1+a_2\cos(\theta_2)+a_3\cos(\theta_2+\theta_3)) \\ \sin(\theta_1)\cos(\theta_2+\theta_3+\theta_4) & -\sin(\theta_1)\sin(\theta_2+\theta_3+\theta_4) & -\cos(\theta_1) & \sin(\theta_1)(a_1+a_2\cos(\theta_2)+a_3\cos(\theta_2+\theta_3)) \\ \sin(\theta_2+\theta_3+\theta_4) & \cos(\theta_2+\theta_3+\theta_4) & 0 & a_2\sin(\theta_2)+a_3\sin(\theta_2+\theta_3)+d_1 \\ 0 & 0 & 0 & 1 \end{bmatrix} \quad (14)$$

Dexterous Hand. Equation 1 will be used to represent all UMDH base configurations. This process suggests that 12 equations can be calculated in 28 operations by reusing similar terms, a 50.9% reduction from the initial 57 Operations. The code, written in *C*, calculates the number of UMDH configurations by controlling four joints via nested FOR loops and then importing the results into MATLAB. In the development phase, functional units were individually designed, tested, and integrated into the VHDL-based forward kinematic processor. The processor, loaded with five constants, conducts a 45-instruction sequence for angle. Top-level testing proved it's functioning, clearing the path for FPGA synthesis.

The subsequent physical implementation and electrical verification focused on a half-sized register file using a Xilinx 4020E FPGA, revealing design size limitations. The ultimate goal of the research is to develop the forward kinematic algorithm for the Utah MIT Dexterous Hand, resulting in a semi-autonomous Forward Kinematic Processor. Despite FPGA constraints, successful implementation and verification were achieved for a resized design with a clock frequency limit of 10.3 MHz. In this case, we should use the highly powerful Altera DE0 NANO SOC FPGA board for physical implementation.

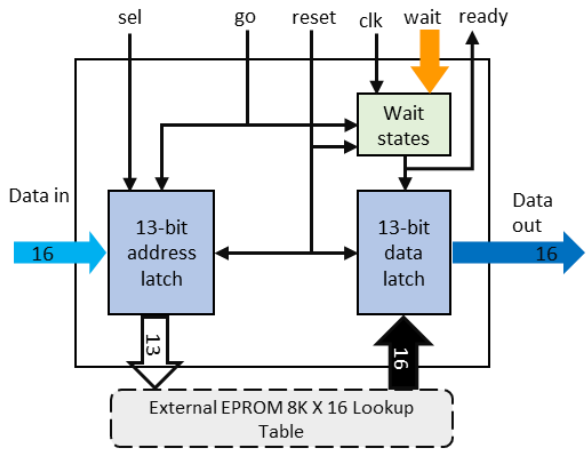

Fig. 11. Architectural diagram of Cosine/Sine unit.

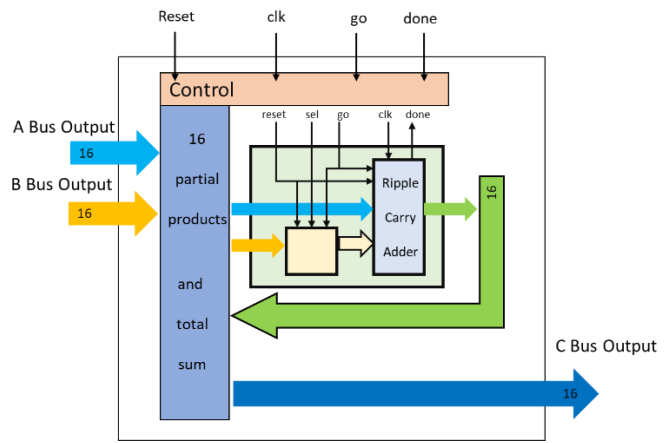

Fig. 13. Architectural diagram of multiplier unit.

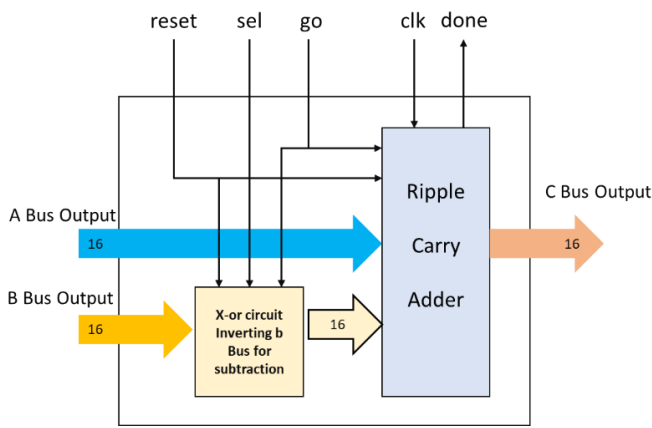

Fig. 12. Architectural diagram of adder/subtractor unit.

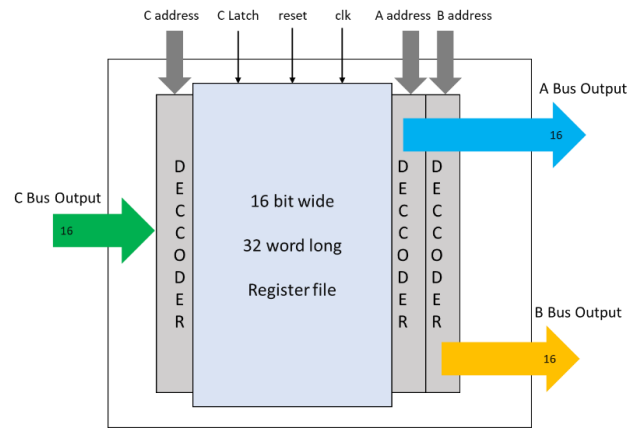

Fig. 14. Architectural diagram of register unit used for FK calculations.

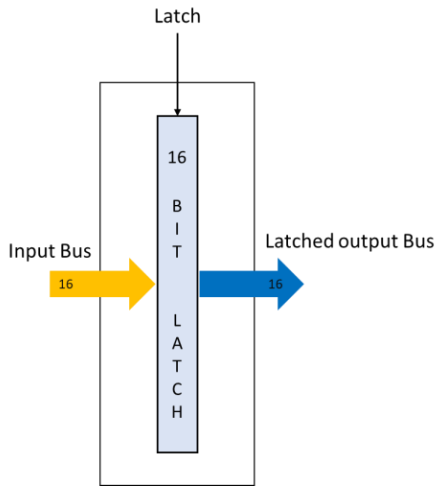

Fig. 15. Architectural diagram of latch unit.

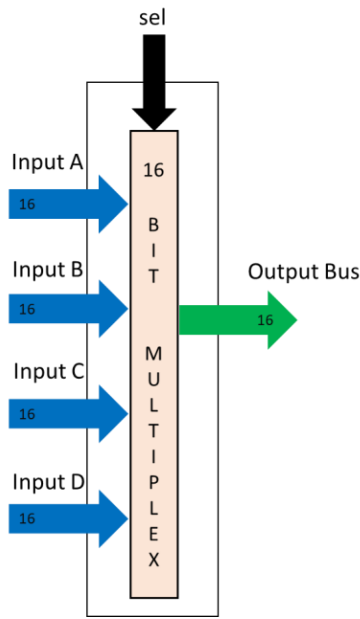

Fig. 16. Architectural diagram of multiplexer unit.

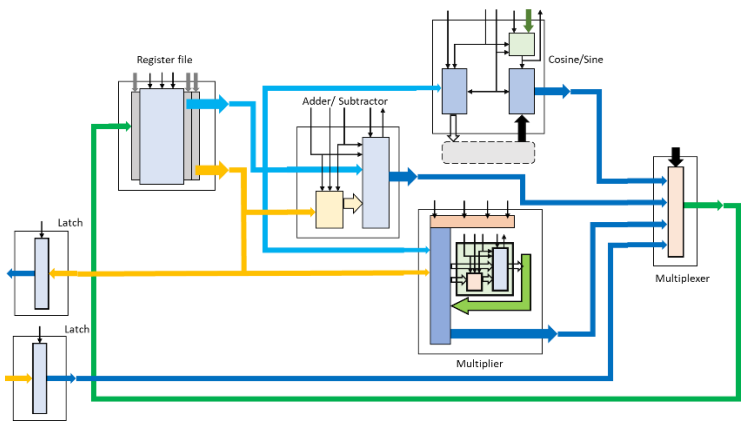

Fig. 17. Architectural diagram of Forward Kinematics processor core.

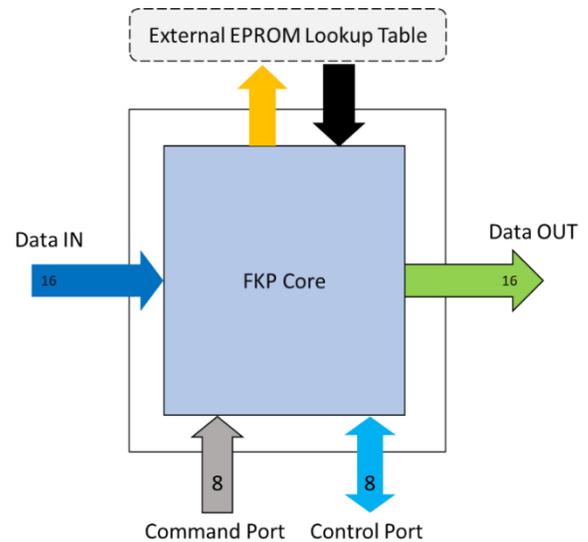

Fig. 18. Architectural diagram of Forward Kinematic (FK) processor system.

## VI. CONCLUSION

In this paper, we have discussed Forward kinematics computation. From the review, we conclude that although the CORDIC processor is highly accurate, it requires to implementation very complex algorithm in the processor. Therefore, it is not suitable beginner level. On the other hand, the method comprising a lookup table is easier to implement and also it has a slightly lower accuracy. Using the lookup table, the computation power required also drops significantly compared to CORDIC architecture. Thus, providing a low-cost and low-computing solution to the CORDIC architecture which will fasten the processing. This survey encapsulates the essence of these innovative pursuits, offering a panoramic view of the strides made in the pursuit of real-time kinematics precision within the realm of robotic manipulators.

## REFERENCES

[1] Zheng, Yili & Liu, Jinhao & Kan, Jiangming. (2012). An Optimal Kinematics Calculation Method for a Multi-DOF Manipulator. *Przeglad Elektrotechniczny*. 88. 320-323.
[2] Lee, C. S. G. and Chen, C. L., "A CORDIC-Based Pipelined Architecture for Direct Kinematic Position Computation" (1987). *Department of Electrical and Computer Engineering Technical Reports.* Paper 553. https://docs.lib.purdue.edu/ecetr/553
[3] Lee and Chen, "A CORDIC-based pipelined architecture for robot direct kinematic position computation," *IEEE 1989 International Conference on Systems Engineering, Fairborn*, OH, USA, 1989, pp. 317-320, doi: 10.1109/ICSYSE.1989.48681.
[4] V. Seshadri, "A real-time VLSI architecture for direct kinematics," Proceedings. 1987 *IEEE International Conference on Robotics and Automation, Raleigh*, NC, USA, 1987, pp. 1116-1120, doi: 10.1109/ROBOT.1987.1087849.
[5] J.. -A. Lee and K. Kim, "Fully-pipelined VLSI architectures for the kinematics of robot arm manipulators," *Eleventh Annual International Phoenix Conference on Computers and Communication [1992 Conference Proceedings]*, Scottsdale, AZ, USA, 1992, pp. 80-86, doi: 10.1109/PCCC.1992.200541.
[6] "FPGA Processor Implementation for The Forward Kinematics of the UMDH", THESIS, Steven M. Parmley, AFIT/GE/ENG/97D-21